\documentclass[runningheads]{llncs}

\pdfoutput=1
\makeatletter
\newcommand{\@chapapp}{\relax}%
\makeatother

\usepackage{graphicx}
\usepackage{amsmath,amssymb} 
\usepackage{color}

\usepackage{tabularx}
\usepackage{booktabs}
\usepackage[inline]{enumitem}
\usepackage[title]{appendix}
\usepackage[absolute]{textpos}
\usepackage{xcolor}

\usepackage[breaklinks=true,colorlinks,allcolors=blue,bookmarks=false]{hyperref} 
\usepackage[capitalize]{cleveref}

\crefname{appsec}{}{}

\newcommand{\ie}{\textit{i.e.}}
\newcommand{\eg}{\textit{e.g.}}

\newcommand{\etal}{et al.\ }
\newcommand*\diff{\mathop{}\!\mathrm{d}}

\DeclareMathOperator*{\argmax}{\arg\!\max}

\definecolor{fsuBlue}{RGB}{135,178,231}

\begin{document}
\title{Information-Theoretic Active Learning for Content-Based Image Retrieval} 

\titlerunning{Information-Theoretic Active Learning}
%
\author{Bj{\"o}rn Barz \and
    Christoph K{\"a}ding \and
    Joachim Denzler}
%
\authorrunning{B. Barz, C. K{\"a}ding, and J. Denzler}
%

\institute{
    Friedrich Schiller University Jena, Ernst-Abbe-Platz 2, 07743 Jena, Germany\\
    \email{bjoern.barz@uni-jena.de}\\
    \url{http://www.inf-cv.uni-jena.de}
 }
\maketitle              

\begin{textblock*}{140mm}(37mm,20mm)
    \textblockcolour{fsuBlue}
    \vspace{2mm}
    \tiny
    \centering
    B. Barz, C. K{\"a}ding, J. Denzler.\\
    Information-Theoretic Active Learning for Content-Based Image Retrieval.\\
    German Conference on Pattern Recognition (GCPR) 2018.\\
    \copyright\ Copyright by Springer. The final publication is available at
    \href{https://link.springer.com/chapter/10.1007\%2F978-3-030-12939-2_45}{link.springer.com}.\\
    \vspace{2mm}
\end{textblock*}

\begin{abstract}
   We propose Information-Theoretic Active Learning (ITAL), a novel batch-mode active learning method for binary classification, and apply it for acquiring meaningful user feedback in the context of content-based image retrieval.
   Instead of combining different heuristics such as uncertainty, diversity, or density, our method is based on maximizing the mutual information between the predicted relevance of the images and the expected user feedback regarding the selected batch.
   We propose suitable approximations to this computationally demanding problem and also integrate an explicit model of user behavior that accounts for possible incorrect labels and unnameable instances.
   Furthermore, our approach does not only take the structure of the data but also the expected model output change caused by the user feedback into account.
   In contrast to other methods, ITAL turns out to be highly flexible and provides state-of-the-art performance across various datasets, such as MIRFLICKR and ImageNet.
   
   \medskip
   
   \textbf{Source code available:} \url{https://github.com/cvjena/ITAL}
\end{abstract}

\section{Introduction}
\label{sec:intro}

\begin{figure}[t]
    \includegraphics[width=\linewidth]{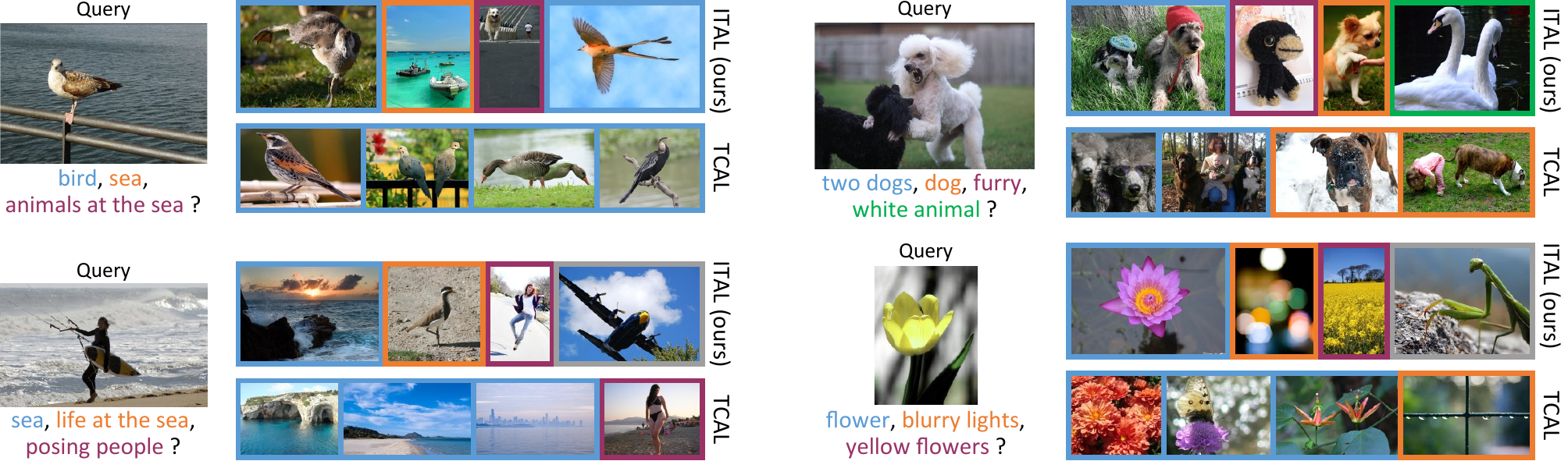}
    \caption{Comparison of candidate batches selected for the second annotation round by ITAL with the selection of TCAL \cite{demir2015tcal} on 4 exemplary queries from the MIRFLICKR \cite{mirflickr} dataset. The border colors correspond to different topics that could be associated with the query. Obviously, ITAL explores much more diverse relevant topics than TCAL.}
    \label{fig:mirflickr-examples}
\end{figure}

For content-based image retrieval (CBIR) \cite{niblack1993qbic,smeulders2000cbir}, it is, in general, not sufficient to just classify the query image or to identify the same object on different images.
Since images encode complex semantic and stylistic information---sometimes more than text can express---a single query is usually insufficient to comprehend the search interest of the user.
A common approach to overcome this issue is enabling the user to provide \textit{relevance feedback} by tagging some retrieval results as relevant or irrelevant \cite{cox2000pichunter,tong2001support,deselaers2008learning}.
This way, however, the user will only be able to give feedback regarding images about which the retrieval system is already very confident.

The effect of relevance feedback could hence be significantly improved when the user is not asked to provide feedback regarding the currently top-scoring results, but for those instances of the dataset that are most informative for the classifier to distinguish between relevant and irrelevant images.
Finding such a set of most informative samples is the objective of \textit{batch-mode active learning (BMAL)} \cite{brinker2003incorporating,guo2008discriminative,giang2014batch,yang2015usdm}, which has recently been explored for CBIR \cite{demir2015tcal,cardoso2017rbmal}.
However, the performance of existing approaches usually varies substantially between datasets, which is not only observable in our experiments, but also in comparative evaluations in the existing literature (\eg, \cite{guo2008discriminative}).

In this work, we propose \textit{Information-Theoretic Active Learning (ITAL)}, a BMAL method for relevance feedback that does not suffer from this instability, but provides state-of-the-art performance across different datasets.
This is demonstrated by a comparison of our approach with a variety of competitor methods on 5 image datasets of very different type and structure.
Our method for selecting unlabeled samples for annotation by the user
\begin{enumerate*}[label=\upshape(\itshape\alph*\upshape)]
    \item implicitly maintains both diversity and informativeness of the candidate images,
    \item employs an explicit model of the user behavior for dealing with the possibility of incorrect annotations and \textit{unnameable instances} \cite{Kaeding15_ALD}, \ie, images which the user cannot classify at all,
    \item takes the model output change caused by the expected user feedback into account,
    \item can easily be parallelized for processing large datasets, and
    \item works with as few as a single initial training sample.
\end{enumerate*}

The user model allows ITAL to compensate for unreliable users, who are likely to make mistakes or to refuse giving feedback.
It acts as an implicit mechanism for controlling the trade-off between redundancy and diversity of the batch of samples selected for annotation.
Because care has to be taken not only that all images in the batch of unlabeled samples selected for annotation are \textit{informative} individually, but that they are also \textit{diverse} compared to each other to avoid unnecessary redundant feedback.
The majority of existing works on BMAL try to achieve this using a combination of several heuristics to simultaneously maximize the \textit{diversity} within the batch and the \textit{uncertainty} of the selected samples or their \textit{density} in the dataset \cite{brinker2003incorporating,zhu2008sud,giang2014batch,demir2015tcal,yang2015usdm,cardoso2017rbmal}.

Our proposed ITAL method, in contrast, aims to maximize the mutual information (MI) between the expected user feedback and the relevance model.
By taking the joint distribution of the predictive relevance of the samples in the batch into account, the MI criterion implicitly maintains diversity without the need for any heuristics or manually tuned linear combinations of different criteria.
Instead, our method does not only take the structure of the data and the current relevance predictions into account, but also considers the expected impact that annotating the selected samples would have on the predicted relevance after updating the model.
This integration of the expected model output change (EMOC) has successfully been used for one-sample-at-a-time active learning \cite{freytag2014EMOC}, but, to the best of our knowledge, not been applied to BMAL yet.

However, computing the expected model output change requires relevance models that can be updated efficiently.
In addition, both the relevance model and the active learning technique should be capable of working with as few training data as a single positive query example provided by the user.
We achieve both by using a Gaussian process (GP) \cite{rasmussen2006gaussian} for classification, which can be fitted to a single training sample and can be updated using a closed-form solution without the need for iterative optimization.
This is in contrast to many other works on active learning, which are based on logistic regression \cite{guo2007optimistic,li2013adaptive} or support vector machines (SVMs) \cite{tong2001support,brinker2003incorporating,giang2014batch,demir2015tcal} as classification technique. 
Moreover, SVMs require a fair amount of both positive and negative initial training data for learning a robust hyperplane.
Thus, such an approach is not feasible for image retrieval.

\Cref{fig:mirflickr-examples} illustrates the advantages of our approach:
While existing methods often select images similar to the query, but with high uncertainty (\eg, only dogs for a dog query or birds for a bird query), ITAL additionally explores the different meanings of the query image.
The query showing a bird in front of the sea could as well refer to images of the sea or to animals at the sea in general.
The dog query, on the other hand, could refer to images showing two dogs, images of dogs in general, or images of white animals.
Finally, the user providing the beach image as query could be interested in images of the coast, of creatures at the beach, or also just in images of people in action without necessarily being at the beach.
All these various options are explored by ITAL, which actively asks the user for the feedback to resolve these ambiguities.

We will briefly review related methods in the following section and explain our ITAL method in detail in \cref{sec:ital}. The experiments mentioned above are presented in \cref{sec:experiments} and \cref{sec:conclusions} concludes this paper.

\section{Related Work}
\label{sec:related-work}

The use of active learning methods is, of course, not limited to information retrieval applications, but also evident in the scenario of manual annotation of large unlabeled datasets:
One would prefer spending money and human effort on labeling the most useful samples instead of outliers.
Thus, active learning has been extensively studied for several years across various application domains, including binary classification \cite{tong2001support,brinker2003incorporating,guo2008discriminative,freytag2013labeling,freytag2014EMOC}, multi-class classification \cite{guo2007optimistic,jain2009pknn,li2013adaptive,Kaeding15_ALD,yang2015usdm}, and regression \cite{guestrin2005near,krause2007nonmyopic,Kaeding18_ALR}.

With regard to batch-mode active learning (BMAL), most existing methods employ some combination of the criteria uncertainty, diversity, and density:
Brinker \cite{brinker2003incorporating} proposes to select samples close to the decision boundary, while enforcing diversity by minimizing the maximum cosine similarity of samples within the batch. 
Similarly, ``Sampling by Uncertainty and Density (SUD)'' \cite{zhu2008sud} selects samples maximizing the product of entropy and average cosine similarity to the nearest neighbors and ``Ranked Batch-mode Active Learning (RBMAL)'' \cite{cardoso2017rbmal} constructs a batch by successively adding samples with high uncertainty and low maximum similarity to any other already selected sample.
``Triple Criteria Active Learning (TCAL)'' \cite{demir2015tcal}, on the other hand, first selects a subset of uncertain samples near the decision boundary, divides them into $k$ clusters, and chooses that sample from each cluster that has the minimum average distance to all other samples in the same cluster.
Following a more complex approach, ``Uncertainty Sampling with Diversity Maximization (USDM)'' \cite{yang2015usdm} finds a trade-off between the individual entropy of the samples in the batch and their diversity by formulating this optimization problem as a quadratic program, whose parameters to be determined are the ranking-scores of the unlabeled samples.

In addition, two works use an information-theoretic approach and are, thus, particularly similar to our method:

Guo \& Greiner \cite{guo2007optimistic} propose to maximize the mutual information between the selected sample and the remaining unlabeled instances, given the already labeled data.
They reduce this objective to the minimization of conditional entropy of the predictive label distribution of the unlabeled samples, given the existing labels and a proxy-label for the selected instance.
With regard to the latter, they make an optimistic guess assuming the label which would minimize mutual information.
If this guess turns out to be wrong, they fall back to uncertainty sampling for the next iteration.

Though the results obtained by this approach called MCMI[min]+MU are convincing, it is computationally demanding and not scalable to real-world scenarios, even though the authors already employed some assumptions to make it more tractable.
In particular, they assume that the conditional entropy of a set of samples can be decomposed as a sum of the entropy of individual samples.
However, this assumption ignores relationships between unlabeled samples and is hence not suitable for a batch-mode scenario.

In our work, we employ different approximations and Gaussian processes to enable the use of mutual information for BMAL.
Using Gaussian processes instead of logistic regression or SVMs also allows us to take the impact of user feedback on the model output into account, since updating a GP does not involve iterative algorithms.

On the other hand, Li \& Guo \cite{li2013adaptive} employ mutual information as a measure for the information density of the unlabeled samples and combine it with the conditional entropy of their individual labels as uncertainty measure.
Similar to our approach, they use a GP to estimate the mutual information, but then employ logistic regression for the actual classification.
Furthermore, their method cannot be applied to a batch-mode scenario and does not scale to large datasets, so that they need to randomly sub-sample the unlabeled data.

Our ITAL method, in contrast, forms a consistent framework, provides a batch-mode, considers the impact of annotations on the model output, and relies solely on the solid theoretical basis of mutual information to implicitly account for uncertainty, density, and diversity.

\section{Information-Theoretic Active Learning}
\label{sec:ital}

We begin with a very general description of the idea behind our ITAL approach and then describe its individual components in more detail.
The implementations of ITAL and the competing methods described in \cref{subsec:competitors} are available as open source at \url{https://github.com/cvjena/ITAL/}.

\subsection{Idea and Ideal Objective}
\label{subsec:idea}

Let $\mathfrak{U} = \{ x_1, \dotsc, x_m \}$ be a set of features of unlabeled samples and $\mathfrak{L} = \{ (x_{m+1}, y_{m+1}), \dotsc, (x_{m+\ell}, y_{m+\ell}) \}$ be a set of features of labeled samples $x_i \in \mathbb{R}^d$ and their labels $y_i \in \{-1,1\}$.
The label $1$ is assigned to relevant and $-1$ to irrelevant samples.
$\mathfrak{X} = \{x_1,\dotsc,x_m,x_{m+1},\dotsc,x_{m+\ell} \}$ denotes the set of all $n=m+\ell$ samples.
In the scenario of content-based image retrieval, $\mathfrak{L}$ usually consists initially of the features of a single relevant sample: the query image provided by the user.
However, queries consisting of multiple and even negative examples are possible as well.

Intuitively, we want to ask the user for relevance feedback for a batch $u \subseteq \mathfrak{U}$ of $k = |u|$ unlabeled samples, whose feedback we expect to be most helpful for classifying the remaining unlabeled instances, \ie, assessing their relevance to the user.
Note that these chosen samples are also often referred to as ``queries'' in the active learning literature.
To avoid confusion caused by this conflicting terminology, we will refer to the query image as ``query'' and to the unlabeled samples chosen for annotation as ``candidates''.

Ideally, the most informative batch $u$ of candidates can be found by maximizing the conditional mutual information $\mathfrak{I}(R,F \mid u)$ between the relevance $R$ of both labeled and unlabeled samples, which is a multivariate random variable over the space $\{-1,1\}^n$ of relevance labels, and the user feedback $F$, being a multivariate random variable over the space $\{-1,0,1\}^n$ of possible feedbacks.
A feedback of $0$ represents the case that the user has not given any feedback for a certain candidate.
This option is a special feature of our approach, which allows the user to omit candidates that cannot be labeled reliably.

Since the size $n$ of the dataset can be huge, this problem is not solvable in practice.
We will show later on how it can be approximated to become tractable.
But for now, let us consider the ideal optimization objective:
\begin{equation}
    \label{eq:objective}
    u = \argmax_{\hat{u} \subseteq \mathfrak{U}} \mathfrak{I}(R,F \mid \hat{u}) \;.
\end{equation}

Writing the mutual information (MI) in terms of entropy reveals the relationship of our approach to uncertainty sampling by maximizing the entropy $H(R \mid u)$ of the candidate batch \cite{li2013adaptive,zhu2008sud,yang2015usdm} or minimizing the conditional entropy $H(R \mid F, u)$ \cite{guo2007optimistic}:
\begin{equation}
    \label{eq:mi-entropy}
    \mathfrak{I}(R,F \mid u) = H(R \mid u) - H(R \mid F, u) \;.
\end{equation}

In contrast to pure uncertainty maximization, we also take into account how the relevance model is expected to change after having obtained the feedback from the user:
To select those samples whose annotation would reduce uncertainty the most, we maximize the difference between the uncertainty $H(R \mid u)$ according to the current relevance model and the uncertainty $H(R \mid F, u)$ after an update of the model with the expected user feedback.

In contrast to existing works \cite{li2013adaptive,zhu2008sud,yang2015usdm}, we do not assume $H(R \mid u)$ to be equal to the sum of individual entropies of the samples, but use their joint distribution to compute the entropy.
Thus, maximizing $H(R \mid u)$ is also a possible novel approach, which will be compared to maximization of MI in the experiments (cf.\ \cref{sec:experiments}).

In more detail, the mutual information can be decomposed into the following components (a derivation is provided in \cref{app:mi-derivation}):
\begin{multline}
    \label{eq:mi-integral}
    \mathfrak{I}(R,F \mid u)
    = \sum_{\substack{r \in \{-1,1\}^n \\ f \in \{-1,0,1\}^n}} \Biggl[ P(R=r \mid u) \cdot P(F=f \mid R=r, u) \\
    \cdot \log\left( \frac{P(R=r \mid F=f, u)}{P(R=r \mid u)} \right) \Biggr] \,.
\end{multline}

\noindent
The individual terms can be interpreted as follows:
\begin{itemize}
    
    \item $P(R \mid u) = P(R)$ is the probability of a certain relevance configuration according to the current relevance model.
    
    \item $P(R \mid F, u)$ is the probability of a certain relevance configuration after updating the relevance model according to the user feedback. Thus, $\frac{P(R \mid F, u)}{P(R \mid u)}$ quantifies the model output change. Compared to other active learning techniques taking model output change into account, \eg, EMOC \cite{freytag2014EMOC}, we do not just consider the change of the predictive mean, but of the joint relevance probability and hence take all parameters of the distributions into account.
    
    \item $P(F \mid R, u)$ is the probability of observing a certain feedback, given that the true relevance of the samples is already known. One might assume that the feedback will always be equal to the true relevance. However, users are not perfect and tend to make mistakes or prefer to avoid difficult samples (so-called \textit{unnameable instances} \cite{Kaeding15_ALD}). Thus, this term corresponds to a \textit{user model} predicting the behavior of the user.
    
\end{itemize}

\noindent
In the following subsections \ref{subsec:rel-model} and \ref{subsec:user-model}, we will first describe our relevance and user model, respectively.
Thereafter, we introduce assumptions to approximate \cref{eq:objective} in the subsections \ref{subsec:approx-mi} and \ref{subsec:approx-greedy}, since finding an optimal set of candidates would require exponential computational effort.

\subsection{Relevance Model}
\label{subsec:rel-model}

We fit a probabilistic regression to the training data $\mathfrak{L}$, \ie, the query images and the images annotated so far, using a Gaussian process \cite[chapter 2]{rasmussen2006gaussian} with an RBF kernel given by the kernel matrix $K \in \mathbb{R}^{n \times n}$ over the entire dataset $\mathfrak{X}$:
\begin{equation}
    \label{eq:rbf-kernel}
    K_{ij} = \sigma_\mathrm{var}^2 \cdot \exp\left(-\frac{\|x_i - x_j\|^2}{2 \cdot \sigma_\mathrm{ls}^2}\right) + \sigma_\mathrm{noise}^2 \cdot \delta_{ij} \;,
\end{equation}
where $\delta_{ij}$ is the Kronecker delta function with $\delta_{ij} = 1 \leftrightarrow i=j$ and zero otherwise, and $\sigma_\mathrm{var}$, $\sigma_\mathrm{ls}$, and $\sigma_\mathrm{noise}$ are the hyper-parameters of the kernel.

The computation of the kernel matrix can be performed off-line in advance and does hence not contribute to the run-time of our active learning method.
However, if time and memory required for computing and storing the kernel are an issue, alternative kernels for efficient large-scale Gaussian process inference \cite{rodner2017large} can be used.

The prediction of the Gaussian process for any finite set of $k$ samples consists of a multivariate normal distribution $\mathcal{N}(\mu, \Sigma)$ over continuous values $\hat{y} \in \mathbb{R}^k$, where $\mu \in \mathbb{R}^k$ is a vector of predictive means of the samples and $\Sigma \in \mathbb{R}^{k \times k}$ is their predictive joint covariance matrix.
Let $p(\hat{y}) = \mathcal{N}(\hat{y} \mid \mu, \Sigma)$ denote the probability density function of such a distribution.

We use this probabilistic label regression for binary classification by considering samples $x_i$ with $\hat{y}_i > 0$ as relevant.
The probability of a given relevance configuration $r \in \{-1,1\}^n$ for the samples in $\mathfrak{X}$ is hence given by
\begin{equation}
    \label{eq:rel-prob}
    P(R=r) = \int_{a_1}^{b_1} \dotsi \int_{a_n}^{b_n} p(y_1,\dotsc,y_n) \diff y_n \dotsm \diff y_1 \;,
\end{equation}
with
\begin{equation}
    \setlength\arraycolsep{0pt}
    a_i = \left\{\begin{array}{rrrr}
    0, & \quad r_i\, &=& \, 1, \\ 
    -\infty, & \quad r_i\, &=& \, -1,
    \end{array}\right. \qquad
    b_i = \left\{\begin{array}{rrrr}
    \infty, & \quad r_i \, &=& \, 1, \\ 
    0, & \quad r_i \, &=& \, -1,
    \end{array}\right.
\end{equation}
for $i = 1, \dotsc, n$.
This is a multivariate normal distribution function, which can be efficiently approximated using numerical methods \cite{genz1992mvndst}.

The posterior probability $P(R \mid F, u)$ can be obtained in the same way after updating the GP with the expected feedback.
For such an update, it is not necessary to re-fit the GP to the extended training data from scratch, which would involve an expensive inversion of the kernel matrix of the training data.
Instead, efficient updates of the inverse of the kernel matrix \cite{luetz2013want} can be performed to obtain updated predictions at a low cost.
This is an advantage of our GP-based approach compared with other methods relying on logistic regression (\eg, \cite{guo2007optimistic,li2013adaptive}), which requires expensive iterative optimization for updating the model.

\subsection{User Model}
\label{subsec:user-model}

We employ a simple, but plausible user model for $P(F \mid R, u)$, which comes along with a slight simplification of the optimization objective in \cref{eq:objective}:
First of all, we assume that if we already know the true relevance $r = [r_1, \dotsc, r_n]^\top$ of all samples, the feedback $f_i$ given by the user for an individual sample $x_i$ is conditionally independent from the feedback provided for the other samples.
More formally:
\begin{equation}
    \label{eq:user-model-independency}
    P(F = f \mid R = r, u) = \prod_{i=1}^{n} P(F_i = f_i \mid R_i = r_i, u) .
\end{equation}

Clearly, if a sample $x_i$ has not been included in the candidate batch $u$, the user cannot give feedback for that sample, \ie, $x_i \notin u \rightarrow f_i = 0$.
Furthermore, we assume that the user will, on average, label a fraction $p_\mathrm{label}$ of the candidate samples.
For each labeled sample, the user is assumed to provide an incorrect label with probability $p_\mathrm{mistake}$.
In summary, this user model can be formalized as follows:
\begingroup
\setlength\arraycolsep{3pt}
\begin{equation}
\label{eq:user-model}
    P(F_i = f_i \mid R_i = r_i, u) = \left\{\begin{array}{crcl}
        0 \,, & x_i \notin u &\wedge & f_i \neq 0, \\ 
        1 \,, & x_i \notin u &\wedge & f_i = 0, \\ 
        1 - p_\mathrm{label} \,, & x_i \in u &\wedge & f_i = 0, \\ 
        p_\mathrm{label} \cdot p_\mathrm{mistake} \,, & x_i \in u &\wedge & f_i \neq r_i, \\
        p_\mathrm{label} \cdot (1 - p_\mathrm{mistake}) \,, & x_i \in u &\wedge & f_i = r_i.
    \end{array}\right.
\end{equation}
\endgroup

The fact that $\left(\exists_{i \in \{1,\dotsc,n\}}: x_i \notin u \wedge f_i \neq 0\right) \rightarrow P(F=f \mid R=r, u) = 0$ allows us to adjust the sum in \cref{eq:mi-integral} to run over only $3^k$ instead of $3^n$ possible feedback vectors, where $k \ll n$ is the batch size and independent from the size $n$ of the dataset.
This is not an approximation, but an advantage of our user model, that decreases the complexity of the problem significantly.

Modeling the user behavior can enable the active learning technique to find a trade-off between learning as fast as possible by asking for feedback for very diverse samples and improving confidence regarding existing knowledge by selecting not extremely diverse, but slightly redundant samples.
The latter can be useful for difficult datasets or tasks, where the user is likely to make mistakes or to refuse to give feedback for a significant number of candidates.
Nevertheless, the assumption of a perfect user, who labels all samples in the batch and never fails, is an interesting special case since it results in a simplification of the MI term from \cref{eq:mi-integral} and can reduce computation time drastically:
\begin{multline}
\label{eq:mi-perfectuser}
    (p_\mathrm{label} = 1 \wedge p_\mathrm{mistake} = 0) \rightarrow \\
    \displaystyle \mathfrak{I}(R,F \mid u) = \displaystyle \sum_{r \in \{-1,1\}^n} \Biggl[ P(R=r \mid u)
    \displaystyle \cdot \log\left( \frac{P(R=r \mid F=r, u)}{P(R=r \mid u)} \right) \Biggr] \,.
\end{multline}

\subsection{Approximation of Mutual Information}
\label{subsec:approx-mi}

Even with the perfect user assumption, evaluating \cref{eq:mi-perfectuser} still involves a summation over $2^n$ possible relevance configurations, which does not scale to large datasets.
To overcome this issue, we employ an approximation based on the assumption, that the probability of observing a certain relevance configuration depends only on the samples in the current candidate batch:
\begin{equation}
    \label{eq:batch-condition-assumption}
    P(R = r \mid u = \{x_{i_1}, \dotsc, x_{i_k}\})
    \;=\; P(R_{i_1} = r_{i_1}, \dotsc, R_{i_k} = r_{i_k}) \;.
\end{equation}
This means that we indeed condition $P(R \mid u)$ on the current batch $u$, though actually $P(R \mid u) = P(R)$ holds for the original problem formulation.

This assumption allows us to restrict the sum in \cref{eq:mi-integral} to $2^k$ instead of $2^n$ possible relevance configurations, leading to the approximate mutual information
\begin{multline}
    \label{eq:mi-approx}
    \tilde{\mathfrak{I}}(R,F \mid u)
    = 2^{n-k} \sum_{\substack{r \in \{-1,1\}^k \\ f \in \{-1,0,1\}^k}} \Biggl[ P(R_u = r) \cdot P(F_u = f \mid R_u = r) \\
    \cdot \log\left( \frac{P(R_u = r \mid F_u = f)}{P(R_u = r)} \right) \Biggr]
\end{multline}
with $u = \{x_{i_1}, \dotsc, x_{i_k}\}$ and, by an abuse of notation, $R_u = [R_{i_1}, \dotsc, R_{i_k}]^\top$ (analogously for $F_u$).
The number of involved summands now does not depend on the size of the dataset anymore, but only on the number $k$ of candidates chosen at each round for annotation.

On the other hand, this assumption also restricts the estimation of the expected model output change, expressed by the term $P(R_u \mid F_u) / P(R_u)$, to the current batch, which is probably the most severe drawback of this approximation.
However, estimating the model output change for the entire dataset would be too expensive and the experiments in \cref{subsec:results} show that our approach can still benefit from the expected model output change.
Future work might explore the option of taking a tractable subset of context into account additionally.

\subsection{Greedy Batch Construction}
\label{subsec:approx-greedy}

Although the computational effort required to calculate the approximate MI given in \cref{eq:mi-approx} is independent from the size of the dataset, finding the exact solution to the optimization problem from \cref{eq:objective} would still require computing the MI for all possible $2^m$ candidate batches $u \subseteq \mathfrak{U}$ of unlabeled samples.
Since we do not want to confront the user with an unlimited number of candidates anyway, we set the batch size to a fixed number $k$, which leaves us with a number of $\binom{m}{k}$ possible candidate batches.

Assessing them all would involve a polynomial number of subsets and is, thus, still too time-consuming in general.
On first sight, one might think that this problem can be solved more efficiently using dynamic programming, but this is unfortunately not an option since the MI is not adequately separable.

Thus, we follow a linear-time greedy approach to approximate the optimal batch by successively adding samples to the batch \cite{guestrin2005near}, taking their relationship to already selected samples into account:
We first select the sample $x_{i_1}$ with maximum $\tilde{\mathfrak{I}}(R, F \mid u = \{x_{i_1}\})$.
The second sample $x_{i_2}$ is chosen to maximize MI together with $x_{i_1}$.
This continues until the batch contains $k$ samples.

At each iteration, the unlabeled samples eligible for being added to the current batch can be treated completely independently from each other, allowing for straightforward parallelization.

\section{Experiments}
\label{sec:experiments}

We demonstrate the performance of our ITAL approach on five image datasets of varying type and structure, described in \cref{subsec:datasets}, and compare it against several existing active learning techniques briefly explained in \cref{subsec:competitors}.
The quantitative results in \cref{subsec:results} show that ITAL is the only method that can provide state-of-the-art performance across all datasets.
Qualitative examples are shown in \cref{fig:mirflickr-examples} and failure cases can be found in \cref{app:failure-cases}.

\subsection{Datasets}
\label{subsec:datasets}

All datasets used in our experiments consist of multiple classes and are divided into a training and a test set.
We define image retrieval tasks for each dataset as follows:
Pick a single random instance from the training set of a certain class as query image and consider all other images belonging to that class as relevant, while instances from other classes are irrelevant.
Batch-mode active learning is performed for 10 successive rounds with a batch-size of $k = 4$ candidates per round and retrieval performance is evaluated after each round by means of average precision on the test set.
This process is repeated multiple times with different random queries for each class and we report the mean average precision (mAP) over all repetitions.

Note that our goal is not to achieve state-of-the-art performance in terms of classification accuracy, but with respect to the active learning objective, \ie, obtaining better performance after fewer feedback rounds.

The smallest dataset used is the \textbf{Butterflies} dataset \cite{butterflies}, comprising 1,500 images of 5 different species of butterflies captured over a period of 100 years.
We use the CNN features provided by the authors and, following their advice, reduce them to 50 dimensions using PCA.
A random stratified subset of 20\% of the dataset is used as test set.

Second, we use the \textbf{USPS} dataset \cite{USPS} consisting of 9,300 gray-scale images of handwritten digits, scanned from envelopes by the U.S.\ Postal Service.
The number of images per class is very unevenly distributed. 
All images have a size of $16 \times 16$ pixels and are used without further feature extraction as 256-dimensional feature vectors.
We use the canonical training-test split provided with the dataset.

As a more real-word use-case, we perform evaluation on the \textbf{13 Natural Scenes} dataset \cite{natural_scenes} and the \textbf{MIRFLICKR-25K} dataset \cite{mirflickr}.
The former consists of more than 3,400 images from 13 categories of natural scenes such as forests, streets, mountains, coasts, offices, or kitchens.
The latter comprises 25,000 images, each assigned to a subset of 14 very general topics such as ``clouds'', ``tree'', ``people'', ``portrait'' etc.
Thus, query images can belong to multiple categories and will be ambiguous.
Asking the user the right questions is hence of great importance.
There are also ``wide-sense annotations'' assigning images to categories if they could be related to a small degree.
If a candidate image is annotated in this way, we consider it as unnameable during our simulation.
For both datasets, we extracted image features from the first fully-connected layer of the VGG-16 convolutional neural network \cite{simonyan2014vgg}, pre-trained for classification on ImageNet, and reduce their dimensionality to 512 using PCA.
Experiments in \cref{app:mirflickr-pca-dim} show that the relative performance of the different methods is not very sensitive w.r.t.\ the dimensionality of the features.
25\% of the natural scenes and 20\% of the MIRFLICKR dataset are used as test set.

Finally, we derive further challenging image retrieval tasks from the \textbf{ImageNet Large Scale Visual Recognition Challenge (ILSVRC)} \cite{ILSVRC15}, which comprises more than 1,2 million images from 1,000 classes.
Following Freytag \etal \cite{freytag2014EMOC}, we obtain binary classification tasks by randomly choosing a single positive and 19 negative classes.
This is repeated 25 times and 10 random queries are chosen for each task, leading to a total of 250 image retrieval scenarios.
We use the bag-of-words (BoW) features provided with ImageNet instead of CNN features, mainly for two reasons: First, the BoW features are public and hence facilitate reproduction. Second, most neural networks are pre-trained on ImageNet, which could bias the evaluation.

The number of random repetitions per class for the natural scenes and the MIRFLICKR dataset has been set to 10 as well, while we use 25 queries per class for USPS and 50 for the butterflies dataset.

The features of all datasets were scaled to be in $[0,1]$.

\subsection{Competitor Methods}
\label{subsec:competitors}

We compare ITAL with a variety of baselines and competing methods, including \textbf{SUD} \cite{zhu2008sud}, \textbf{TCAL} \cite{demir2015tcal}, \textbf{RBMAL} \cite{cardoso2017rbmal}, and the method of Brinker \cite{brinker2003incorporating} referred to as ``\textbf{border\_div}'' in the following.
All these native BMAL methods have been described in \cref{sec:related-work}.

In addition, we evaluate the following successful one-by-one active learning techniques in the BMAL scenario by selecting the $k$ samples with the highest selection scores:
Uncertainty sampling for SVM active learning by choosing samples close to the decision boundary (\textbf{border}) \cite{tong2001support}, uncertainty sampling for Gaussian processes (\textbf{unc}) \cite{kapoor2007active}, where uncertainty is defined as the ratio between absolute predictive mean and predictive standard deviation, and sample selection by maximizing the expected model output change (\textbf{EMOC}) \cite{freytag2014EMOC}.

All methods have to compete against the baselines of \textbf{random} selection, selecting the \textbf{topscoring} samples with maximum predictive mean, resembling the standard retrieval scenario \cite{ayache2007evaluation}, and variance sampling (\textbf{var}) by maximizing the difference of the sum of variances and the sum of covariances in the batch.

Finally, we also investigate maximizing the \textbf{joint entropy} $H(R_{i_1}, \dotsc, R_{i_k}) = - \sum_{r \in \{-1,1\}^k} P(R_u=r) \cdot \log(P(R_u=r))$ of candidate batch $u = \{x_{i_1},\dotsc,x_{i_k}\}$.
Being a component of our ITAL method, this is also a novel approach, but lacks the model output change term and the user model (cf.\ \cref{eq:mi-entropy}).

We also tried applying \textbf{USDM} \cite{yang2015usdm}, \textbf{MCMI[min]} \cite{guo2007optimistic} and \textbf{AdaptAL} \cite{li2013adaptive}, especially since the latter two also maximize a mutual information criterion.
However, all these methods scale so badly to datasets of realistic size, that they could not be applied in practice.
AdaptAL, for example, would require 14 hours for composing a single batch on MIRFLICKR, which is clearly intractable.
Our ITAL method, in contrast, can handle this dataset with less than a minute per batch.
These three competitors could, thus, only applied to USPS, MIRFLICKR, and ImageNet by randomly sub-sampling 1000 candidates to choose from, as suggested by Li \etal \cite{li2013adaptive}.
This usually leads to a degradation of performance, as can be seen from the results reported in \cref{app:adaptal}.

\subsection{Hyper-parameters}
\label{subsec:hyper-params}

The hyper-parameters of the RBF kernel, \ie, $\sigma_\mathrm{ls}$, $\sigma_\mathrm{var}$, and $\sigma_\mathrm{noise}$ (cf. \cref{subsec:rel-model}), potentially have a large impact on the performance of the active learning methods.
However, the overall goal is to eventually obtain a classifier that performs as well as possible.
Therefore, we determine the optimal kernel hyper-parameters for each dataset using tenfold cross-validation on the training set and alternating optimization to maximize mean average precision.
This optimization aims only for good classification performance independent of the active learning method being used and the same hyper-parameters are used for all methods.

With regard to the hyper-parameters of the user model used by ITAL, we employ the perfect user assumption for being comparable to competing methods that do not model the user.
An experiment evaluating the effect of different user model parameters is presented in \cref{subsec:exp-user-model}.

In case that other methods have further hyper-para\-me\-ters, we use the default values provided by their authors.

\subsection{Results}
\label{subsec:results}

\begin{figure}[t]
    \includegraphics[width=\linewidth]{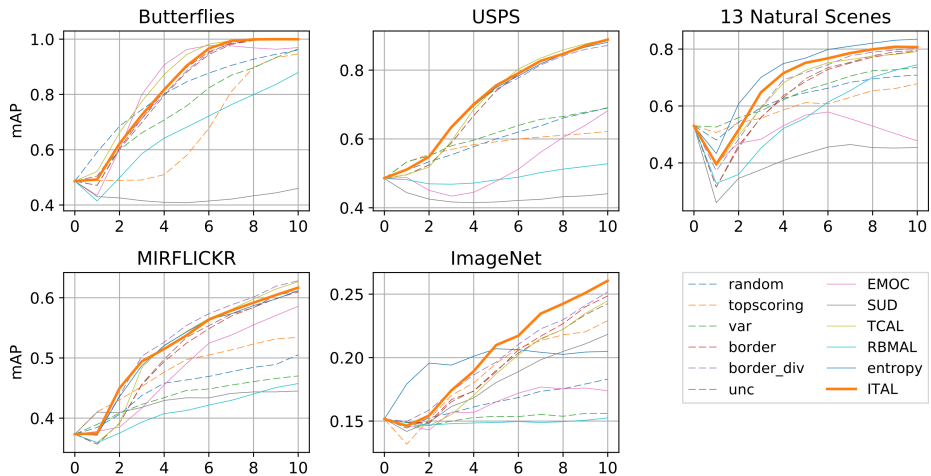}
    \caption{Comparison of retrieval performance after different numbers of feedback rounds for various active learning methods. The thick, orange line corresponds to our proposed method. Figure is best viewed in color.}
    \label{fig:performance}
\end{figure}

\Cref{fig:performance} depicts the average precision obtained on average after 10 feedback rounds using the different BMAL methods.
On the Butterflies dataset, ITAL obtains perfect performance after the least number of feedback rounds.
TCAL and border\_div perform similar to ITAL on USPS, but ITAL learns faster at the beginning, which is important in interactive image retrieval scenarios.
While sampling candidates based on batch entropy behaves almost identical to ITAL on Butterflies, USPS, and MIRFLICKR, it is slightly superior on the Natural Scenes dataset, but fails to improve after more than 4 rounds of feedback on the ImageNet benchmark, where ITAL is clearly superior to all competitor methods.
This indicates that taking the effect of the expected user feedback on the model output change into account is of great benefit for datasets as diverse as ImageNet.

Since it is often desirable to compare different methods by means of a single value, we report the area under the learning curves (AULC) in \cref{table:auc}, divided by the number of feedback rounds so that the best possible value is always 1.0.
In all cases, our method is among the top performers, achieving the best of all results in 3 out of 5 cases.
The improvement over the second-best method on 13 Natural Scenes and ImageNet is significant on a level of \textless1\% and on a level of 7\% on USPS, according to Student's paired t-test.

\begin{table}[t]
    \caption{Area under the learning curves from \cref{fig:performance}. The numbers in parentheses indicate the position in the ranking of all methods. The best value in each column is set in bold face, while the second-best and third-best values are underlined.}
    \label{table:auc}
    \begin{scriptsize}
        \setlength\tabcolsep{1pt}
        \begin{tabularx}{\linewidth}{lXccXccXccXccXccXr}
            \toprule
            && \multicolumn{14}{c}{Area under Learning Curve (AULC)} && \\
            \cmidrule{3-16}
            Method && \multicolumn{2}{c}{Butterflies} && \multicolumn{2}{c}{USPS} && \multicolumn{2}{c}{Nat.\ Scenes} && \multicolumn{2}{c}{MIRFLICKR} && \multicolumn{2}{c}{ImageNet} && Avg. Rank \\
            \midrule
            random		&& 0.7316	& (8)	&& 0.5416	& (8)	&& 0.5687	& (8)	&& 0.4099	& (9)	&& 0.1494	& (9)	&& 8.4 \\
            topscoring	&& 0.5991	& (11)	&& 0.5289	& (9)	&& 0.5419	& (9)	&& 0.4358	& (7)	&& 0.1708	& (7)	&& 8.6 \\
            var			&& 0.6800	& (9)	&& 0.5550	& (7)	&& 0.5831	& (5)	&& 0.3957	& (10)	&& 0.1383	& (11)	&& 8.4 \\
            border		&& 0.7434	& (6)	&& 0.6393	& (5)	&& 0.5775	& (7)	&& 0.4559	& (6)	&& 0.1743	& (4)	&& 5.6 \\
            border\_div	&& 0.7456	& (5)	&& \underline{0.6465}	& (3)	&& \underline{0.6031}	& (3)	&& \textbf{0.4795}	& (1)	&& \underline{0.1791}	& (3)	&& \underline{3.0} \\
            unc			&& 0.7373	& (7)	&& 0.6391	& (6)	&& 0.5793	& (6)	&& 0.4585	& (5)	&& 0.1725	& (5)	&& 5.8 \\
            EMOC		&& \underline{0.7561}	& (2)	&& 0.4723	& (10)	&& 0.4654	& (11)	&& 0.4357	& (8)	&& 0.1483	& (10)	&& 8.2 \\
            SUD			&& 0.3887	& (12)	&& 0.3903	& (12)	&& 0.3766	& (12)	&& 0.3883	& (11)	&& 0.1626	& (8)	&& 11.0 \\
            TCAL		&& \textbf{0.7720}	& (1)	&& 0.6459	& (4)	&& 0.6016	& (4)	&& 0.4688 & (4)	&& 0.1708	& (6)	&& 3.8 \\
            RBMAL		&& 0.6023	& (10)	&& 0.4457	& (11)	&& 0.5046	& (10)	&& 0.3732	& (12)	&& 0.1356	& (12)	&& 11.0 \\
            \midrule[.03em]
            entropy (ours)	&& \underline{0.7512}	& (3)	&& \underline{0.6484}	& (2)	&& \textbf{0.6547}	& (1)	&& \underline{0.4703}	& (3)	&& \underline{0.1793}	& (2)	&& \underline{2.2} \\
            ITAL (ours)	&& 0.7511	& (4)	&& \textbf{0.6522}	& (1)	&& \underline{0.6233}	& (2)	&& \underline{0.4731}	& (2)	&& \textbf{0.1841}	& (1)	&& \textbf{2.0} \\
            \bottomrule
        \end{tabularx}
    \end{scriptsize}
\end{table}

The performance of the competing methods, on the other hand, varies significantly across datasets.
ITAL, in contrast, is not affected by this issue and provides state-of-the-art performance independent from the characteristics of the data.
To make this more visible, we construct a ranking of the tested methods for each dataset and report the average rank in \cref{table:auc} as well.
ITAL achieves the best average rank and can thus be considered most universally applicable.

This is of high importance because, in an active learning scenario, labeled data is usually not available before performing the active learning.
Thus, adaptation of the AL method or selection of a suitable one depending on the dataset is difficult.
A widely applicable method such as ITAL is hence very desirable.

\subsection{Effect of the User Model}
\label{subsec:exp-user-model}

To evaluate the effect of the user model integrated into ITAL, we simulated several types of users behaviors on the 13 Natural Scenes dataset:
\begin{enumerate*}[label=\itshape\alph*\upshape)]
    \item\label{item:aggressive-user} an aggressive user annotating all images but assigning a wrong label in 50\% of the cases,
    \item\label{item:conservative-user} a conservative user who always provides correct labels but only annotates 25\% of the images on average, and
    \item\label{item:realistic-user} a blend of both, labeling 50\% of the candidate images on average and having a 25\% chance of making an incorrect annotation.
\end{enumerate*}

The same active learning methods as in the previous sections are applied and the parameters $p_\mathrm{label}$ and $p_\mathrm{mistake}$ of ITAL are set accordingly.

The results presented in \cref{app:imperfect-users} show that the user model helps ITAL to make faster improvements than with the perfect user model.
A possible reason for this effect is that the parameters of the user model control an implicit trade-off between diversity and redundancy used by ITAL:
For an imperfect user, selecting samples for annotation more redundantly can help to reduce the impact of wrong annotations.
However, ITAL performs reasonably well even with the perfect user assumption.
This could hence be used to speed-up ITAL noticeably with only a minor loss of performance.

\section{Conclusions}
\label{sec:conclusions}

We have proposed information-theoretic active learning (ITAL), a novel batch-mode active learning technique for binary classification, and applied it successfully to image retrieval with relevance feedback.
Based on the idea of finding a subset of unlabeled samples that maximizes the mutual information between the relevance model and the expected user feedback, we propose suitable models and approximations to make this NP-hard problem tractable in practice.
ITAL does not need to rely on manually tuned combinations of different heuristics, as many other works on batch-mode active learning do, but implicitly trades off uncertainty against diversity by taking the joint relevance distribution of the instances in the dataset into account.

Our method also features an explicit user model that enables it to deal with unnameable instances and the possibility of incorrect annotations.
This has been demonstrated to be beneficial in the case of unreliable users.

We evaluated our method on five image datasets and found that it provides state-of-the-art performance across datasets, while many competitors perform well on certain datasets only.
Moreover, ITAL outperforms existing techniques on the ImageNet dataset, which we attribute to its ability of taking the effect of the expected user feedback on the model output change into account.

\section*{Acknowledgements}

This work was supported by the German Research Foundation as part of the
priority programme ``Volunteered Geographic Information: Interpretation,
Visualisation and Social Computing'' (SPP 1894, contract number DE 735/11-1).

\bibliographystyle{splncs04}
\bibliography{references}

\begin{appendices}
\renewcommand{\thesection}{\appendixname~\Alph{section}}
\crefalias{section}{appsec}

\section{Performance of MCMI[min] and AdaptAL}
\label{app:adaptal}

Since MCMI[min] \cite{guo2007optimistic} and AdaptAL \cite{li2013adaptive} also maximize a mutual information criterion and are, thus, similar to our method, we also tried to apply those methods to our benchmark datasets.
Even though we replaced the expensive logistic regression with Gaussian process inference for being comparable to our method, they could only be applied to the Butterflies and 13 Natural Scenes dataset within reasonable time. For the remaining 3 datasets, we randomly sub-sampled 1000 candidates from the entire dataset, as suggested by \cite{li2013adaptive}.

\begin{table}
    \caption{Comparison of ITAL with MCMI[min] and AdaptAL in terms of AULC.}
    \label{tbl:mi-comp}
    \setlength\tabcolsep{6pt}
    \begin{tabularx}{\linewidth}{Xccccc}
        \toprule
        Method & Butterflies & USPS & Nat.\ Scenes & MIRFLICKR & ImageNet \\
        \midrule
        random & 0.7316 & 0.5416 & 0.5687 & 0.4099 & 0.1494 \\
        MCMI[min] & 0.6846 & 0.5293 & 0.4554 & 0.4087 & 0.1413 \\
        AdaptAL & \textbf{0.7716} & 0.6487 & 0.6424 & 0.4643 & 0.1746 \\
        \midrule
        entropy (ours) & 0.7512 & 0.6484 & \textbf{0.6547} & 0.4703 & 0.1793 \\
        ITAL (ours) & 0.7511 & \textbf{0.6522} & 0.6233 & \textbf{0.4731} & \textbf{0.1841} \\
        \bottomrule
    \end{tabularx}
\end{table}

The results in \cref{tbl:mi-comp} show that MCMI[min] does not work well in a batch-mode scenario and performs worse than random.

AdaptAL, on the other hand, is the top performer on the Butterflies dataset and the second-best method on Natural Scenes, directly behind our batch-entropy approach.
These are the two datasets where it could be applied in reasonable time on the entire dataset.
The sub-sampling that is necessary on the remaining three datasets, however, negatively impacts performance, especially on ImageNet.

To the best of our knowledge, our method is the first one that makes an information-theoretic approach to batch-mode active learning applicable in realistic scenarios without sub-sampling the dataset.

\newpage

\section{Simulation of Imperfect Users}
\label{app:imperfect-users}

As described in section 4.5 of the paper, we have investigated the effect of three different extreme user behavior models on the performance of the tested BMAL methods.
With regard to our approach, we have evaluated both ITAL with the user model parameters $p_\mathrm{label}$ and $p_\mathrm{mistake}$ set according to the simulated user and ITAL with the perfect user assumption, which is faster.

We have selected batches of 4 images for annotation at each round.

\begin{figure}
    \includegraphics[width=\linewidth]{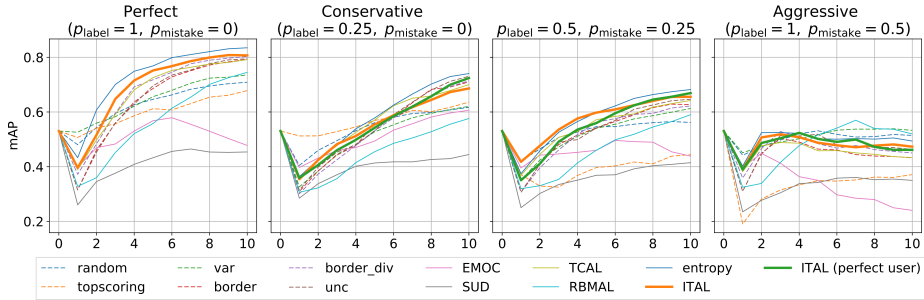}
    \caption{Comparison of different user behavior models on Natural Scenes.}
    \label{fig:user-behaviors}
\end{figure}

As expected, all methods suffer from imperfect user feedback compared to a perfect user.
While an adequate user model helps ITAL to learn faster during the first rounds, the difference is small enough to justify the use of the perfect user assumption even if it is not true in order to gain a significant speed-up.
The case of overly aggressive but error-prone users obviously cannot be handled by the active learning method alone, but also requires adequate handling of such scenarios by the classifier.

\newpage

\section{Sensitivity of Results regarding Feature Dimensionality}
\label{app:mirflickr-pca-dim}

\begin{figure}
    \centering
    \includegraphics[width=.75\linewidth]{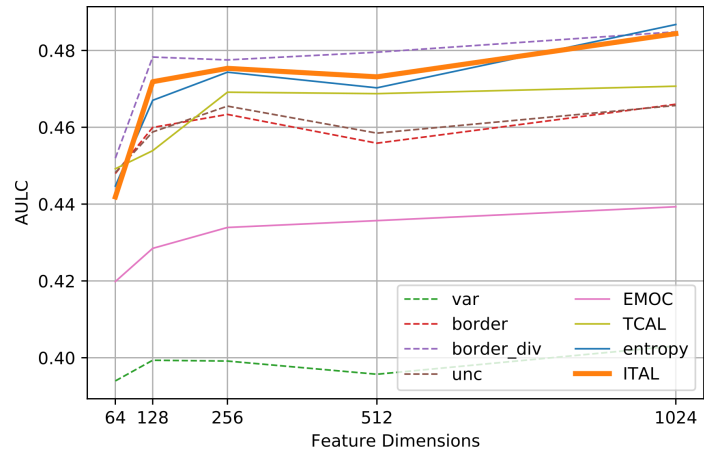}
    \caption{Area under Learning Curve (AULC) of various BMAL methods on MIRFLICKR with varying feature dimensionality.}
    \label{fig:mirflickr-pca-dim}
\end{figure}

To assess to which extent the results presented in the paper are affected by certain transformations applied to the features, we experimented with different dimensionalities of the feature space on the MIRFLICKR dataset.
To this end, we have applied PCA to the features extracted from the first fully-connected layer of VGG16, which comprise 4096 dimensions, and projected them onto spaces with 64, 128, 256, 512, and 1024 features.
Experiments with all BMAL methods have been conducted on those features for 10 rounds of user feedback and the area under the learning curve (AULC) for the various dimensionalities is reported in \cref{fig:mirflickr-pca-dim}.

The results show that the relative performance of the different methods compared to each other is largely insensitive to the number of features.
The performance of ITAL is stable up to as few as 128 dimensions, while some other methods such as TCAL and EMOC already degrade after reducing the number of features to less than 256.
When using 1024 features, ITAL is even able to catch up to border\_div, which is the best performing method on this particular dataset.
However, we have used 512 features for our experiments in the paper due to the increased computational cost incurred by higher-dimensional feature spaces.

\newpage

\section{Examples for Failure Cases}
\label{app:failure-cases}

\begin{figure}[h]
    \centering
    \includegraphics[width=\linewidth]{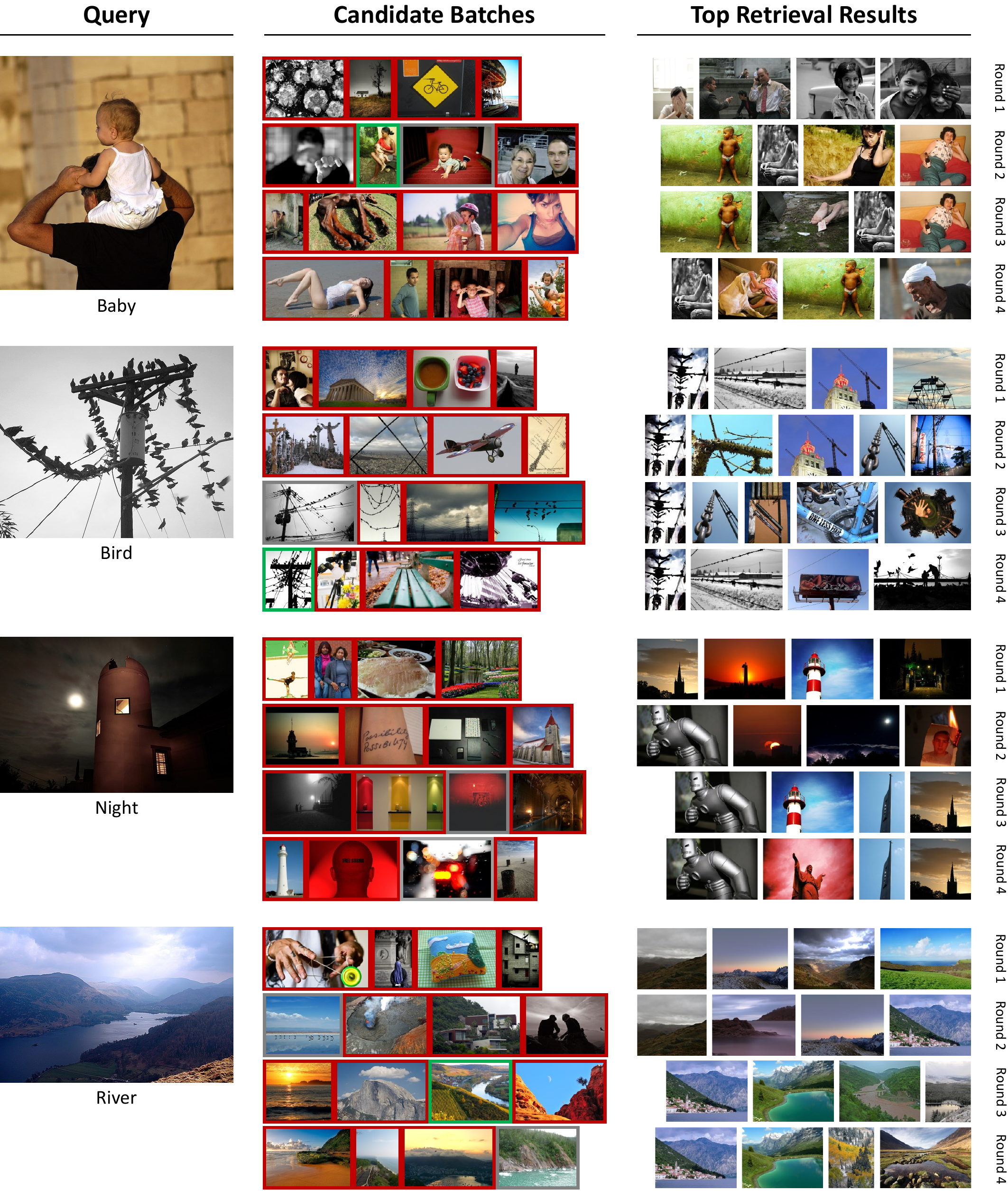}
    \caption{Four queries from MIRFLICKR where ITAL performed worst.}
    \label{fig:failure-cases}
\end{figure}

To analyze the possible shortcomings of our method, we have picked four queries from the MIRFLICKR dataset where ITAL had the worst AULC score.
These are depicted in \cref{fig:failure-cases}, along with the candidate images selected for annotation over 4 rounds of feedback and the top results retrieved by the relevance model after each round.

The first query could be interpreted in multiple ways:
The user could be searching for images of people, of babies, or of adults with babies.
All these options are covered by the candidate images selected by ITAL.
Only one of those image shows a baby alone, which is the actual search objective in this example.
That image, however, has not been annotated confidently as showing a baby in the MIRFLICKR dataset, so that it remains unnameable here.

The second query shows a swarm of birds on a power pole, but the simulated user actually searches for birds.
The features used in our experiment are apparently not sufficient to capture the semantics of this image well enough for recognizing that it is about birds.
Thus, the selected candidates do not contain any image of a bird in a different scene and the classifier cannot abstract away from power poles.

The ``night'' query, on the other hand, is again an example of erroneous annotations in the dataset:
Several images of night scenes have been selected as candidates, but have been annotated either as unnameable or even as irrelevant.

Finally, the last query image shows a river and the candidates are actually quite suitable to identify whether the user is more interested in mountain scenes, water scenes, river scenes, or natural scenes in general.
However, either the features or the small number of annotated images seem to be insufficient in this case for distinguishing between rivers and other bodies of water.

\clearpage

\section{Derivation of Eq. (3)}
\label{app:mi-derivation}

Plugging in the definitions of entropy and conditional entropy into the definition of mutual information given in eq. (2) leads to the following:

\begin{multline*}
    \mathfrak{I}(R,F \mid u) =
    - \Biggl[ \sum_{r \in \{-1,1\}^n} P(R=r \mid u) \cdot \log P(R=r \mid u)  \Biggr] + \\
    \Biggl[ \sum_{\substack{r \in \{-1,1\}^n \\ f \in \{-1,0,1\}^n}} P(F=f \mid u) \cdot P(R=r \mid F=f, u) \cdot \log P(R=r \mid F=f, u) \Biggr] \;.
\end{multline*}

\noindent
Expressing $P(R=r \mid u)$ in the first sum as the marginalization
\begin{equation*}
    P(R=r \mid u) = \sum_{f \in \{-1,0,1\}^n} P(F=f \mid u) \cdot P(R=r \mid F=f, u)
\end{equation*}
allows us to merge the two sums:

\begin{multline*}
    \mathfrak{I}(R,F \mid u) =
    \sum_{\substack{r \in \{-1,1\}^n \\ f \in \{-1,0,1\}^n}} \Biggl[
        P(F=f \mid u) \cdot P(R=r \mid F=f, u) \\
        \cdot \log \left( \frac{P(R=r \mid F=f, u)}{P(R=r \mid u)} \right) \;.
    \Biggr]
\end{multline*}

\noindent
Using Bayes' Theorem we can substitute
\begin{equation*}
    P(F=f \mid u) \cdot P(R=r \mid F=f, u) = P(R=r \mid u) \cdot P(F=f \mid R=r, u) \;,
\end{equation*}
finally leading to eq. (3) from the main paper.

\end{appendices}

\end{document}